\documentclass[conference]{IEEEtran}
\usepackage{cite}
\usepackage{amsmath,amssymb,amsfonts}
\usepackage{algorithmic}
\usepackage{graphicx}
\usepackage{textcomp}
\usepackage{xcolor}
\def\BibTeX{{\rm B\kern-.05em{\sc i\kern-.025em b}\kern-.08em
    T\kern-.1667em\lower.7ex\hbox{E}\kern-.125emX}}
\begin{document}

\title{Neuro-Symbolic AI: An Emerging Class of AI Workloads and their Characterization\\
}

\author{\IEEEauthorblockN{\textsuperscript{} Zachary Susskind, Bryce Arden, Lizy K. John}
\IEEEauthorblockA{\textit{Department of Electrical and Computer Engineering} \\
\textit{The University of Texas at Austin}\\
Austin, Texas \\
zsusskind, arden.bryce, ljohn@utexas.edu}
\and
\IEEEauthorblockN{\textsuperscript{} Patrick Stockton, Eugene B. John}
\IEEEauthorblockA{\textit{Department of Electrical and Computer Engineering} \\
\textit{The University of Texas at San Antonio}\\
San Antonio, Texas \\
patrick.stockton, eugene.john@utsa.edu}
}

\maketitle

\begin{abstract}
Neuro-symbolic artificial intelligence is a novel area of AI research which seeks to combine traditional rules-based AI approaches with modern deep learning techniques. Neuro-symbolic models have already demonstrated the capability to outperform state-of-the-art deep learning models in domains such as image and video reasoning. They have also been shown to obtain high accuracy with significantly less training data than traditional models. Due to the recency of the field's emergence and relative sparsity of published results, the performance characteristics of these models are not well understood. In this paper, we describe and analyze the performance characteristics of three recent neuro-symbolic models.
We find that symbolic models have less potential parallelism than traditional neural models due to complex control flow and low-operational-intensity operations, such as scalar multiplication and tensor addition. However, the neural aspect of computation dominates the symbolic part in cases where they are clearly separable. We also find that data movement poses a potential bottleneck, as it does in many ML workloads.
\end{abstract}

\begin{IEEEkeywords}
Neuro-Symbolic, Machine Learning, Performance, Inference
\end{IEEEkeywords}

\section{Introduction}
\label{introduction}

Conventional neural networks, based on Deep Learning (DL), have proven to be effective in solving problems in many domains.
Recently, there has been an increase in the diversity of neural models, including those composed of multiple independently-trained submodels, and those which integrate symbolic reasoning concepts (neuro-symbolic models).
At the same time, there is growing interest in heterogeneous architectures. These architectures pose the challenge of efficiently mapping problems to different resources. Even in a relatively simple heterogeneous design, such as a CPU-GPU system, it is necessary to determine which tasks should be mapped to which device.
In order to facilitate this mapping, it is important to understand emerging AI applications.\par


Neural networks have traditionally been grey-box systems. It's trivial to observe the internal \textit{physical} structure of a network, since a model's topology is explicitly defined as part of hyperparameter selection. However, mapping the high-dimensional, abstract features which neural networks manipulate to concrete concepts which humans can understand has proven to be difficult. Such mappings are highly desirable for safety-critical fields such as medicine and self-driving cars, where we must be certain a model is learning the desired dataset features rather than over-fitting to data artifacts \cite{MONTAVON20181}. Networks composed of independent submodels have an advantage here, since each submodel was trained for a specific purpose. Knowing the purpose of each submodel gives us some information about the internal state of the top-level model.\par

Neuro-Symbolic AI (NSAI) is another emerging AI domain that combines deep learning for feature extraction and rules-based "intuition" for manipulating those features. Rules-based, or symbolic, approaches dominated the field of AI until the 1980s \cite{GARNELO201917}. Symbolic models had several advantages: they required only a few input samples, generalized well to new problems, and their internal functionality was conceptually simple when compared to DL models. At the same time, they required substantial hand-tuning, which made them difficult to create for complex problems. A far larger issue was that they simply weren't very accurate: in 1973, the entire field of AI was summarised as "increasingly disappointing"\cite{lighthill}, and by the 1980s, research had spiraled into what became known as the "AI winter"\cite{hendler2008avoiding}.\par

\begin{figure}[hbtp]
\centerline{\includegraphics[width = 0.8\columnwidth]{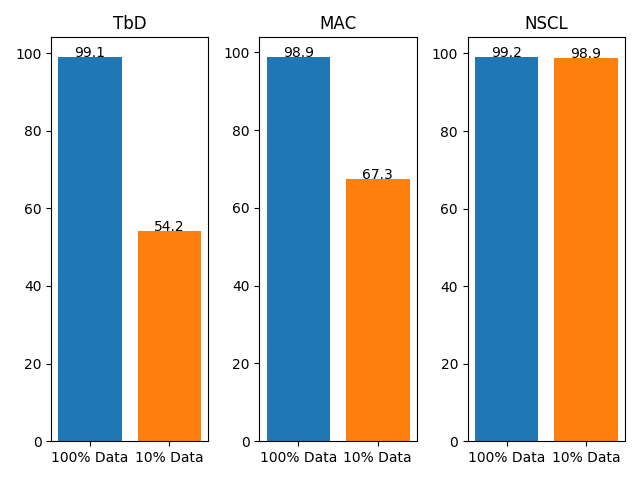}}
\caption{The neuro-symbolic model (NSCL\cite{mao2019neurosymbolic}) achieves good accuracy even when only trained with 10\% of the data (CLEVR dataset) whereas the accuracy of neural models (TbD\cite{tbd_model} and MAC\cite{mac_model}) deteriorates when the training set is small. Data is drawn from \cite{mao2019neurosymbolic}.} 
\label{fig:partial_training}
\end{figure}

NSAI is a hybrid between DL and symbolic approaches which attempts to capture the strengths of both fields. Deep learning has proven singularly successful in extracting complex features from data in tasks such as object detection and natural language processing. At the same time, symbolic AI is good for formalizing human-like reasoning. The objective of NSAI is extract features from data using DL approaches, then manipulate these features using symbolic approaches. Neuro-symbolic models have outperformed pure DL models on image and video question answering tasks, and have proven to converge far quicker, with as little as 1/10 of the training data needed for accuracy. Figure \ref{fig:partial_training} contrasts the NSCL, a neuro-symbolic model for image reasoning~\cite{mao2019neurosymbolic}, with two other pure-DL models~(TbD\cite{tbd_model} and MAC\cite{mac_model}). While all three models obtain similar final accuracies with 100\% of the training data, the neuro-symbolic model dramatically outperforms the other two when restricted to training on only 10\% of the total data (CLEVR dataset) \cite{mao2019neurosymbolic}.
Therefore, neuro-symbolic models are a good choice for scenarios where the availability of training data is limited, or when training time is prohibitive.
\par

In this paper, we analyze the inference performance characteristics of three separate neuro-symbolic models. Two of the models, the Neuro-Symbolic Concept Learner (NSCL) and Neuro-Symbolic Dynamic Reasoning (NS-DR), are composed of multiple independent "submodels", which extract problem features before passing them as input to a final symbolic submodel. The third, Neural Logic Machines (NLM), is an end-to-end model without independent submodels. The NLM model uses an object’s relations, properties, quantifiers, and logic connectives in order to accomplish the task of generalization. We provide an overview of all three models in Table \ref{tab:model_overview}. \par

\begin{table*}[t]
\centering
\caption{Overview of the three neuro-symbolic models discussed in this paper}
\begin{tabular}{|l|l|p{0.4\textwidth}|p{0.35\textwidth}|} 
    \hline
    Workload & Source & Description & Submodels \\ [0.5ex] 
    \hline
    NSCL\cite{mao2019neurosymbolic} & MIT/IBM & A model that learns visual concepts, words, and semantic parsing of sentences without explicit supervision on any of them. Model learns by simply looking at images and reading paired questions and answers. The specific MIT model builds an object-based scene representation and translates sentences into executable, symbolic programs. Uses CLEVR dataset. &
    \begin{itemize} 
    \item Image Parser (Object detection/masking)
    \item Question Parser (Natural Language Processing)
    \item Symbolic Executor
    \end{itemize}\\
    \hline
    NS-DR\cite{yi2020clevrer} & MIT/IBM & A neuro-symbolic model for the CLEVRER dataset. CLEVRER includes 300,000 questions and answers related to the action depicted in the videos. The questions fall into four different categories: Descriptive: e.g. “What is the material of the last object to collide with the cyan cylinder? Explanatory: e.g. “What is responsible for the collision between rubber and metal cylinders?” Predictive: e.g. “What will happen next? Counterfactual: e.g. “What will happen without the cyan cylinder?” &
    \begin{itemize} 
    \item Video Frame Parser (Object detection/masking)
    \item Question Parser (Natural Language Processing)
    \item Dynamics Predictor (Learned physics)
    \item Symbolic Executor
    \end{itemize}\\
    \hline
    NLM\cite{neural_logic_machines} & Google & A neuro-symbolic architecture for both inductive learning and logic reasoning. Exploits the power of both neural networks as function approximators, and logic programming as a symbolic processor for objects with properties, relations, logic connectives, and quantifiers. After being trained on small-scale tasks (such as sorting short arrays), NLM models can recover lifted rules, and generalize to large-scale tasks (such as sorting longer arrays than they were trained on) & 
    \begin{itemize}
    \item No Submodels
    \end{itemize}\\
    \hline
\end{tabular}
\label{tab:model_overview}
\end{table*}

The remainder of this paper is organized as follows: Section \ref{related_work} outlines the space of neuro-symbolic learning, including the progressive improvements to neuro-symbolic models. Section \ref{model_overview} describes the three models that we are analyzing in this paper in detail. Section \ref{methodology} describes our methodology for analyzing the performance of these models, based on classifying activity into eight distinct categories. Section \ref{results} provides the results of our analysis with breakdowns for each submodel component. Section \ref{analysis} provides our takeaways on the behavior and potential opportunities for acceleration of neuro-symbolic models. Finally, in Section \ref{conclusion}, we summarize our findings and propose future work. We also provide direct links to the repositories of the models cited in this paper as an appendix.\par

\section{Related Work}
\label{related_work}

A large portion of the research in NSAI to date has been spearheaded by a collaborative effort between MIT and IBM research groups~\cite{mao2019neurosymbolic}, \cite{han2020visual}. In 2019, the Neuro-Symbolic Concept Learner proved the viability of combining symbolic AI with deep learning techniques by parsing input questions and scenes into symbolic programs \cite{mao2019neurosymbolic}. The Concept Learner demonstrated a novel technique for parsing input scenes and questions into a semantic program, and introduced new techniques for training this parsing at the same time as the symbolic execution engine. The Concept Learner model was followed the next year by the Visual Concept-Metaconcept Learning (VCML) model, which used embeddings to learn object properties with less supervision \cite{han2020visual}. The VCML model introduced the joint learning of concepts and metaconcepts (e.g. the notion that two values describe the same property of objects), and the autonomous learning of new concepts (e.g. color, shape, etc.). The VMCL obtained near-perfect accuracy on the CLEVR image reasoning dataset. In early 2020, Chuang Gen et al. introduced the CLEVER dataset, which pushed the boundaries of dynamic reasoning by using videos instead of images, requiring models to learn both physical and causal relationships \cite{yi2020clevrer}. The same paper also introduced the Neuro-Symbolic Dynamic Reasoning (NS-DR) model, which achieved state-of-the-art performance on the CLEVRER dataset, dramatically outperforming traditional, non-symbolic approaches to video reasoning. In early 2021, the same MIT researchers released the Dynamic Concept Learner (DCL) \cite{zfchen2021iclr}, which achieved a new state-of-the-art for the CLEVRER dataset by adding learned features for object behavior over time and using new training techniques. The DCL has also been shown to generalize well to new datasets for video question answering \cite{zfchen2021iclr}.\par

The neuro-symbolic Neural Logic Machine (NLM) research was conducted by a collaboration between Google Inc., ByteDance Inc., and Tsinghua University \cite{neural_logic_machines}. The resulting NLM architecture provides a state-of-the-art method for solving general application tasks such as array sorting, critical path finding, and more complex tasks such as Blocks World \cite{blocks_world}. Blocks World is a classic symbolic reasoning problem where the model is given a set number of blocks and logical rules. Using the provided generalized rules, the model will need to perform the available logical actions to achieve the desired target result from the randomized starting layout. Following the neuro-symbolic work from Jiayuan Mao et al. introduced the concept learner for interpreting scenes, words, and sentences from natural supervision (NS-CL) \cite{mao2018the}. This NS-CL model utilized similar neuro-symbolic reasoning modules to build an object-based scene for the system to infer. The visual reasoning of the NS-CL model generalized tasks using new object attributes and properties with scaling rule sets which is supportive of the NLM work.

As mentioned in the introduction, the NS-DR and NSCL models are composed of independent submodels. Many of these are neural models similar to those already deployed in data centers, covering tasks such  as natural language and image processing\cite{low_latency_rnns} \cite{lightweight_mask_rcnn}. While the approximate performance profiles for some of these submodels is already known, the renewed interest in symbolic AI is relatively recent, and combining symbolic execution with neural networks is quite novel. NLM takes a different approach, integrating its symbolic and neural components, and does not contain any submodels. The field is still growing and evolving rapidly; to the best of our knowledge, there is no prior work which analyzes neuro-symbolic workloads.

\section{Model Overview}
\label{model_overview}

\begin{table*}[t]
\centering
\caption{Overview of the submodel components of the NSCL and NS-DR}
\begin{tabular}{|l|l|l|p{0.5\textwidth}|} 
    \hline
    Submodel & Origin & Type & Description \\ [0.5ex] 
    \hline
    Image/Video Frame Parser & Detectron2\cite{wu2019detectron2} & Neural (CNN) &
    Extracts object features (positions, shapes, materials, sizes, and colors) from still images or video frames.\\
    \hline
    Question Parser & OpenNMT\cite{klein2017opennmt} & Neural (Transformer) &
    Translates natural-language input questions into a purpose-designed "language" of tokens. Tokens may describe the properties of objects, or describe filtering and querying operations to be performed on the set of object features.\\
    \hline
    Dynamics Predictor & PropNet\cite{li2018propagation} & Neural &
    Learned physics engine for modeling and predicting collisions between objects. Is capable of accurately modeling complex collisions involving more than two objects, as well as operating with partial information (allowing objects to enter and leave the scene).\\
    \hline
    Symbolic Executor & NSCL\cite{mao2019neurosymbolic} & Quasi-Symbolic &
    Fixed-function model which sequentially applies parsed tokens to extracted image features to answer input questions. Unlike a true symbolic model, approximates non-differentiable functions using probabilistic approaches and softmax, enabling backpropagation.\\ 
    \hline
    Symbolic Executor & NS-DR\cite{yi2020clevrer} & Symbolic &
    Fixed-function model which applies parsed tokens to both extracted video frame features and PropNet predicted features (collisions). A true symbolic model, it largely performs arithmetic and boolean operations as well as table lookups and queries.\\
    \hline
\end{tabular}
\label{tab:submodels}
\end{table*}

\subsection{Neuro-Symbolic Concept Learner}
\begin{figure}[hbtp]
\centerline{\includegraphics[width = 0.8\columnwidth]{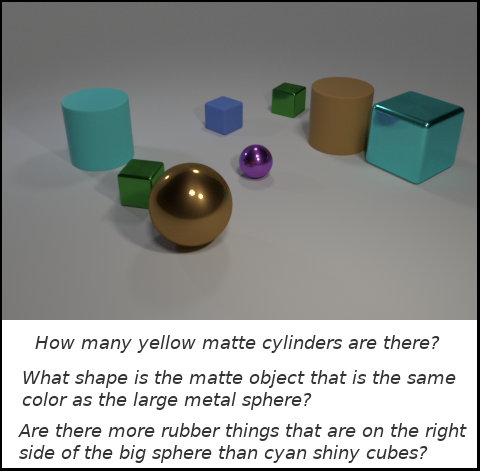}}
\caption{A sample from the CLEVR image reasoning dataset with several accompanying questions. Questions explore the properties of and relations between objects in the scene.}
\label{fig:clevr_example}
\end{figure}

The Neuro-Symbolic Concept Learner (NSCL) was designed for the CLEVR dataset \cite{clevr_datset}. CLEVR is a dataset for "image reasoning": images are presented to the model, along with a set of related questions, and the model's outputs are the answers to these questions.
Image samples in CLEVR contain cubes, cylinders, and spheres with different sizes, colors, and materials. An example of a scene in the CLEVR dataset, with several of its accompanying questions, is shown in Figure \ref{fig:clevr_example}.\par

As described in Table \ref{tab:model_overview}, the NSCL is internally composed of three submodels. We describe each of these submodels in more detail here.

\subsubsection{Image Parser}
The objective of the image parser is to generate object "masks": pixel-accurate regions with annotated colors, shapes, and materials. This is accomplished using a Mask R-CNN model, which constructs object segmentation masks and classifications in parallel \cite{he2018mask}. While the addition of object mask generation makes its structure somewhat more complex than a traditional convolutional neural network, both branches of computation are internally convolutional.\par

The implementation of Mask R-CNN originally used for the NSCL proved challenging to run on modern hardware. Not only was no pretrained model provided, but the software libraries required were obsolete and not compatible with modern CUDA versions. Therefore, we decided to use a more recent, pretrained Mask R-CNN model provided by Facebook's Detectron2 \cite{wu2019detectron2}. This also had the advantage of giving us access to the profiling tools built into more recent versions of the PyTorch framework.

\subsubsection{Question Parser}
Questions in the CLEVR dataset are in the form of natural language, which presents the challenge of translating them into a form usable by the model. The authors of the NSCL accomplished this by defining a domain-specific language, including verbs, such as "Filter" or "Intersect", and concepts, such as "Blue" or "Left". This effectively converts the problem of question parsing into neural machine translation (NMT). A bidirectional GRU\cite{cho-etal-2014-learning} was used to accomplish this task for the original NSDR. However, source code, a pretrained model, or any other information is not available for this model. As such, we chose to profile a small, modern pretrained Transformer-based NMT model provided by Harvard's OpenNMT toolkit\cite{klein2017opennmt}. This provides similar advantages to the Detectron2 image model: more recent library versions and support for modern DL profiling tools. It also provides a more realistic insight into what deployment of this model would look like in a modern datacenter environment.

\subsubsection{Symbolic Program Executor}
The symbolic program executor converts extracted image features and question tokenizations into predictions. Formally, this model is "quasi-symbolic": unlike a true symbolic AI model, which performs logical and boolean operations on data, the executor approximates non-differentiable functions in a probabilistic manner. The output of the NSCL is a vector in which the $i$'th element, $\text{Mask}_{i} \in [0,\,1]$, represents the probability of the corresponding object in the scene being in the output set. This representation is useful for two reasons. First, it allows for multiple correct answers, which increases the complexity of the questions the model can answer. Second, it makes the symbolic step fully reversible, which allows back-propagation to occur. This approach avoids the limitations of true Boolean circuits, which are in general irreversible. In turn, this reversibility allows for the learning of concept embedding vectors: representations of object features which allow concepts (such as color, shape, and material) to be learned without explicit labels\cite{mao2019neurosymbolic}.

\subsection{Neuro-Symbolic Dynamic Reasoning}
The Neuro-Symbolic Dynamic Reasoning (NS-DR) model is a neuro-symbolic model for the CLEVRER video reasoning dataset. CLEVRER builds on CLEVR by replacing images with videos. Objects in CLEVRER may enter or exit the scene throughout the course of a (5 second, 25 frame) video, potentially colliding and rebounding with other objects. Using videos instead of images adds a new domain to the questions in CLEVRER: that of causality. As described in Table \ref{tab:model_overview}, causal questions may range from the descriptive ("Which objects collided?") to more complex causal relations ("What caused the objects to collide? Would the objects still have collided if ..."). Existing neural models were shown to perform poorly on all but the simple descriptive questions in CLEVRER \cite{yi2020clevrer}.\par

In order to account for causal relations, the NS-DR introduces a new submodel: a neural dynamics predictor, which is essentially a learned physics engine. The introduction of the dynamics predictor brings the model up to a total of four independent submodels. We discuss the structure below.

\subsubsection{Video Frame Parser}
The video frame parser treats each frame of a video separately, using the same Mask R-CNN approach as the NSCL. Thus, for each input video, inference with this model must be run 25 times.

\subsubsection{Question Parser}
The original NS-DR model used a more modern NMT model than the NSCL: Seq2Seq\cite{bahdanau2016neural}, which was demonstrated to be more accurate on long inputs than prior models. Once again, we opted to replace this model with the OpenNMT transformer model.

\subsubsection{Dynamics Predictor}
The dynamics predictor, PropNet \cite{li2018propagation}, is a learned physics engine that can represent complex collisions between objects. PropNet improves on prior work in the domain by accurately modeling propagation of force through multiple objects (such as in a Newton's cradle), and operating correctly in the presence of partial information (where not all objects are visible). Functional correctness with partial information is crucial, since all videos in CLEVRER are taken from a fixed camera angle, where objects are allowed to enter and leave the scene. Dynamics prediction provides the positions, trajectories, and collisions between objects for the NS-DR model. The results of the dynamics predictor are augmented with the properties identified by the video frame parser to provide a complete record of what occurred during the input video.

\subsubsection{Symbolic Program Executor}
The program executor of the NS-DR is a true symbolic model: unlike the NSCL, it uses non-differentiable operations to make predictions. The disadvantage of non-differentiable operations is that it is not possible to back-propagate error; therefore, this model can not learn concept embeddings, so concepts must be learned directly by the frame parser via supervised training. This is an entirely fixed-function model with no learned component; internally, it behaves much like a programming language interpreter, using tokens to apply filter and reduction operations to the extracted video features.\par

\begin{figure}[htbp]
\centerline{\includegraphics[width = 0.9\columnwidth]{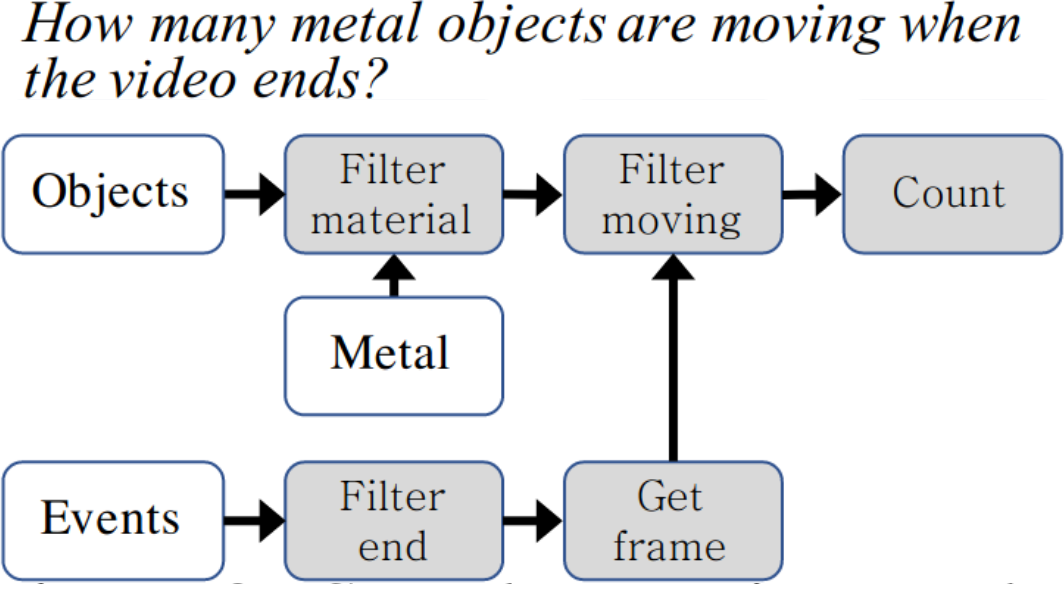}}
\caption{An example of the tokenized representation of a question in the CLEVRER dataset. "Noun/Adjective" tokens - features - have white backgrounds, while "verb" tokens - actions - are shaded. Arrows show the dependencies for token processing\cite{yi2020clevrer}.}
\label{fig:symbolic_program}
\end{figure}

The NS-DR program executor is single-threaded and CPU-only. The sequential nature of its processing does not expose any obvious opportunities for the sorts of coarse-grain parallelism typical of DL workloads: in general, processing the $n$th token of a sequence will require the result of processing the $(n-1)$th token. An example symbolic program, shown in Figure \ref{fig:symbolic_program}, demonstrates how the symbolic program executor uses tokens to filter and perform basic arithmetic on the features extracted by the other submodels.

\subsection{Neural Logic Machines}
\begin{figure*}[t]
\centerline{\includegraphics[width = \textwidth]{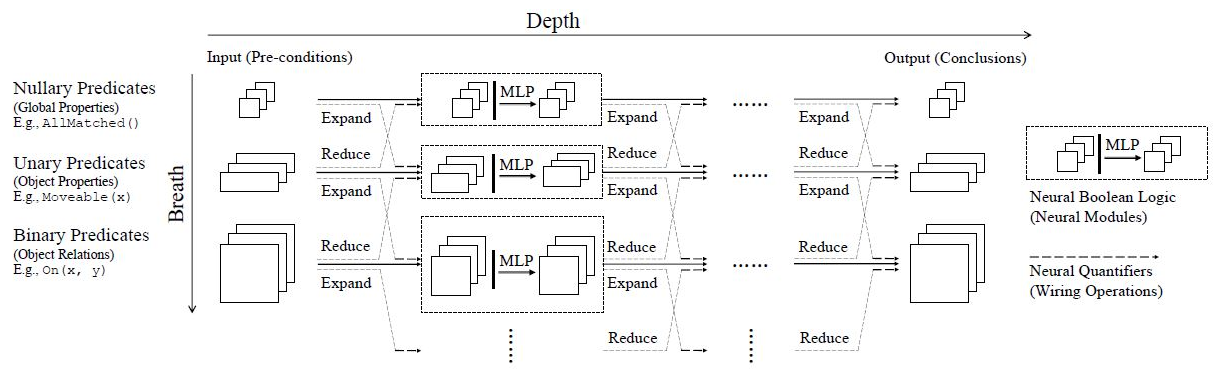}}
\caption{An illustration of the NLM framework showing object properties and object relations as inputs, the concluding outputs of the objects properties and relations, and the internal logical structure. Image originally from \cite{neural_logic_machines}.}
\label{fig:nlm_framework}
\end{figure*}

The Neural Logic Machine (NLM) is a neuro-symbolic architecture that applies both inductive learning and logic reasoning to an object's relations, properties, quantifiers, and logic connectives in order to accomplish the task of generalization. Unlike most traditional networks, NLMs are able to achieve perfect generalization in various tasks such as family tree decisions, sorting of arrays, finding the shortest path between points, or playing the Blocks World task. The NLM architecture provides the ability to solve historically challenging tasks that traditional neural network architectures struggle to complete. The challenge of generalization of tasks from small scale to large scale are proven to be overcome using NLM. Figure \ref{fig:nlm_framework} shows an illustration for the NLM framework.

\section{Methodology}
\label{methodology}

We used function-level profiling to capture statistics such as runtimes, invocation counts and tensor sizes. We then developed a post-processing tool to partition the profiling results into eight dominant categories of operations.

\subsection{Experimental Methodology}
The characterization of all GPU workloads was performed using the built-in PyTorch profiler (\texttt{torch.autograd.profiler}). This profiler collects CPU and GPU runtimes at a per-function granularity, differentiating between "total" runtime (including sub-functions) and "self" runtime (excluding sub-functions). In most cases, this profiling introduced some overhead, and in a few cases, it appeared to capture its own post-run analysis as overhead. We have compensated for this to the extent possible by comparing runs with and without the profiler enabled.\par

Since the symbolic executor for the NS-DR is a fixed-function, CPU-only model, it can not be effectively analyzed using the PyTorch profiler. Instead, we used Python's cProfile module to get per-function execution times. The cProfile module has several limitations when compared to the PyTorch profiler - for instance, it does not profile GPU activity, and does not report input tensor sizes. However, these limitations have no impact on scalar, CPU-only code.\par

Data collection was performed on a system with two Intel Xeon E5-2698 v3 processors (totaling 32 cores and 64 threads), 128GB of DDR4 memory at 2133MHz, and two NVIDIA Tesla M40s. However, most models were only single-threaded, and were only able to make efficient use of a single GPU for inference.\par


\subsection{Workload Taxonomy}
Since simply reporting per-function profiling results would require us to discuss an overwhelming number of individual functions, we devised a function classification scheme to partition the data into eight dominant categories of operations. We drew inspiration for this classification from the Berkeley "seven dwarfs"\cite{berkeley_dwarfs}: a mid-2000s effort to identify what were then the most crucial kernels for parallel computing. In fact, some of our categories are identical to Berkeley "dwarfs", but we identify several others that also appear highly relevant to the workloads discussed in this paper and/or machine learning more generally. Given the wide diversity of models in modern AI research, we do not claim that our classification scheme is universally applicable. However, we do hope to show that workload classification schemes are valuable for exploring the differences between models.\par

\begin{figure}[htbp]
\centerline{\includegraphics[width = \columnwidth]{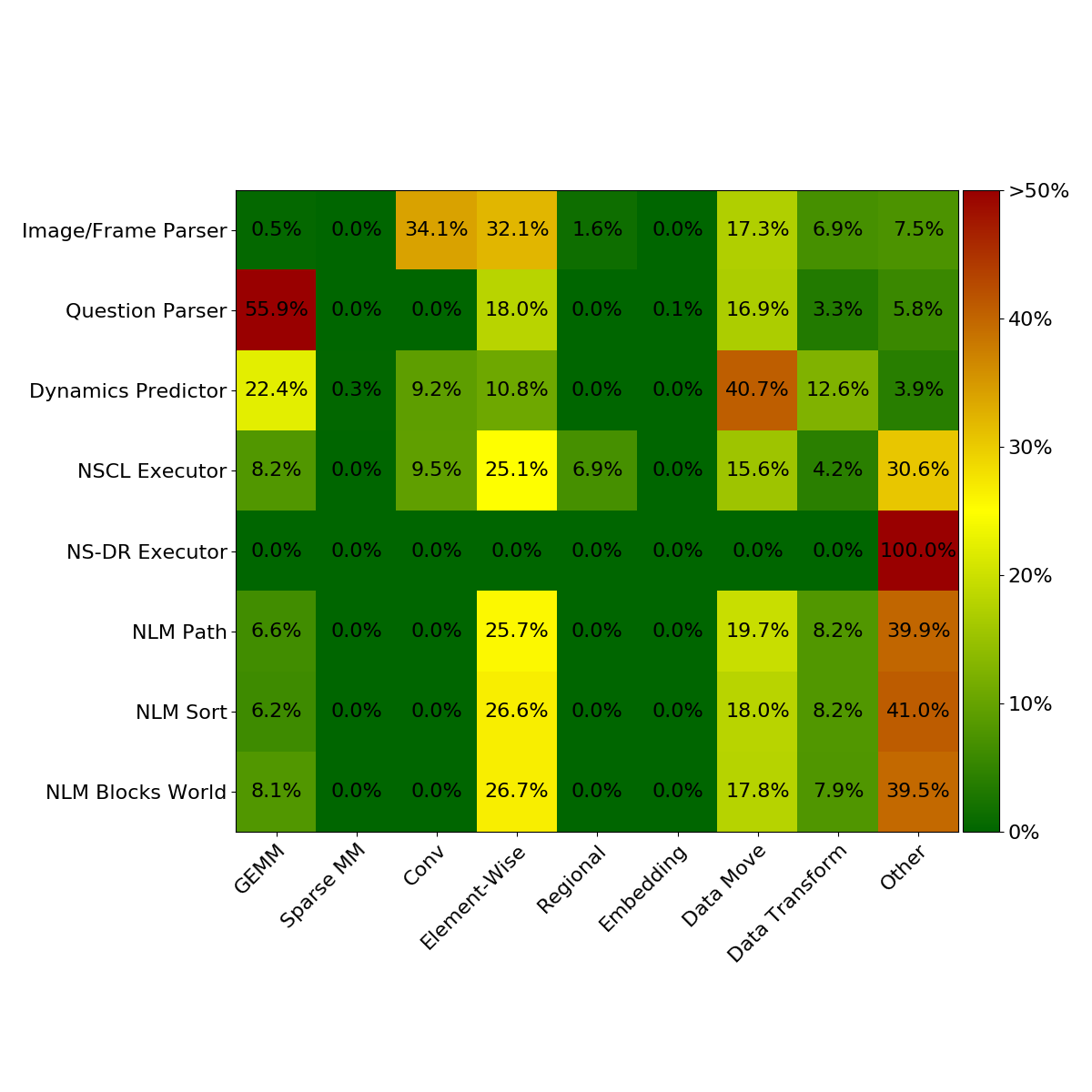}}
\caption{Heatmap showing the proportion of each category of operation in each model. Note that the neural models have different dominant components; the NSCL executor and NLM models are dominated by element-wise operations. See appendix for links to the model repositories.}
\label{fig:heatmap}
\end{figure}

\subsubsection{Dense Matrix Multiplication}
Fast, efficient dense matrix multiplication (GEMM) remains a critical requirement for large DL models. Fully-connected layers in neural networks use GEMM as their primary mathematical operation, often with very large input matrices.\par
Multiplication of large, dense matrices is very computationally intensive: the work to multiply a $m \times k$ matrix with a $k \times m$ matrix $W = \mathcal{O}(mnk)$. At the same time, data intensity only grows quadratically with input size: given the two input matrices and $m \times n$ output matrix, the data intensity $Q = \mathcal{O}(mk + kn + mn) = \mathcal{O}(\max(mk,\,kn,\,mn))$. This gives an operational intensity of:
\[
    I = \frac{W}{Q} = \frac{\mathcal{O}(mkn)}{\mathcal{O}(\max(mk,\,kn,\,mn))} = \mathcal{O}(\min(m,k,n))
\]
When $m$, $k$, and $n$ are all large, operational intensity can be very high. Since there are no internal dependencies in matrix multiplication (the multiply-and-add operations can be performed in any order), the multiplication of large, roughly-square matrices is highly parallelizable. The challenge emerges when any one of the dimensions is small relative to the other two; in this case, the operational intensity approaches $\mathcal{O}(1)$, requiring highly efficient data movement to avoid becoming memory-bound. Such "tall-and-skinny" matrices are difficult to process efficiently on GPUs \cite{Rivera_2021}. While operational intensity can sometimes be addressed by processing multiple inputs simultaneously via batching, this may not be an option for latency-sensitive inference operations where input must be processed as soon as it is received. An extreme case of tall-and-skinny GEMM is the multiplication of a matrix by a vector, as an $n$-element vector can be viewed as an $n \times 1$ matrix.

\subsubsection{Sparse Matrix Multiplication}
Sparse matrices are those in which the values of only some elements are specified; all other elements are assumed to be some constant value, typically 0. There are numerous ways to implement sparse matrices; in general, there is a tradeoff between the generality of the sparsity (how much structure is assumed) and how easy it is to implement in hardware\cite{structured_sparsity}. Sparse matrix multiplication requires efficient mechanisms to perform lookups into the tables of non-zero values.

\subsubsection{Convolution}
Convolutions are another class of common, computationally intensive operation. Unlike in GEMM, one of the inputs, the filter, is multiplied with a set of submatrices of the other input. In theory, this should lead to excellent operational intensity, since filter weights are reused for each individual multiplication, and weights from the other matrix are reused by overlapping submatrices. In practice, convolutions have proven difficult to parallelize, and are frequently implemented using the im2col algorithm, which reduces the problem to matrix multiplication. Performing im2col requires substantial data movement and duplication, worsening the performance and memory footprint of convolution. Recent efforts have focused on performing convolution "natively", without invoking im2col \cite{fast_conv}. The more complex memory access patterns of these approaches have their own considerations for efficient implementation. Whether im2col is used or not, convolution presents a workload distinct from simple GEMM.

\subsubsection{Element-wise Tensor Operations}
Many operations in deep learning are applied uniformly to all elements in a tensor. These include activation, normalization, tensor addition, Hadamard products, tensor-scalar operations, and relational operations. All of these operations are "embarrassingly parallel" in that they can be applied to every element of a tensor simultaneously, but are challenging in that they have poor operational intensity; each element of the input is only operated on once.

\subsubsection{Regional Operations}
Some operations act on spatially local regions of tensors. The best-known example of a regional operation is pooling, which reduces the size of a tensor in one or more dimensions by performing some reduction operation (such as max or average) regionally. This is not the only class of regional operation; other examples include non-maximum suppression and region-of-interest alignment in object detection networks. These operations are distinct from element-wise operations in that they operate on potentially overlapping regions rather than single elements and thus have more complex access patterns; they are distinct from convolution in that they operate on only a single tensor and typically involve less computation.

\subsubsection{Embedding Lookups}
Embeddings are a way of transforming data indices or high-dimensional one-hot vectors into low-dimensional learned vectors. In practice, during inference, embeddings function as lookup tables. The irregular memory access patterns of embedding lookups and the large sizes of embedding tables make acceleration a challenge. Traditionally, embedding lookups are latency-sensitive, memory-bound operations for inference \cite{embedding_acceleration}.

\subsubsection{Data Movement}
Many types of operations require substantial data movement but little or no computation. This primarily consists of host-device and device-host transfers; we also include operations such as tensor duplication and assignment.

\subsubsection{Data Transformation}
The last category of operations we identify is data transformation: operations which reshape or subsample data. This includes matrix transposes, tensor reordering, and masked selection. We also include coalescing in this category, which is a process in which duplicate entries for the same coordinates in a sparse matrix are eliminated by summing their associated values \cite{pytorch_sparse_doc}.

\section{Results}
\label{results}

\begin{table*}[t]
\centering
\caption{Runtimes and runtime breakdowns for single inputs to the models discussed in this paper.}
\begin{tabular}{|c|c|c|c|c|c|c|c|c|c|c|} 
    \hline
    Model & GEMM & Sparse MM & Conv & Element-Wise & Regional & Embedding & Data Move & Data Transform & Other & Total \\
    \hline
    Image/Frame Parser & 0.19ms & 0ms & 11.8ms & 11.1ms & 0.54ms & 0ms & 6.0ms & 2.4ms & 2.6ms & 34.6ms \\
    \hline
    Question Parser & 166ms & 0ms & 0ms & 53.5ms & 0ms & 0.27ms & 50.1ms & 9.9ms & 17.3ms & 297ms \\
    \hline
    Dynamics Predictor & 715ms & 9.9ms & 294ms & 345ms & 0ms & 0ms & 1300ms & 403ms & 125ms & 3200ms \\
    \hline
    NSCL Executor & 39.9us & 0us & 46.4us & 122.4us & 33.5us & 0.0us & 76.0us & 20.5us & 149.5us & 488.3us \\
    \hline
    NS-DR Executor & N/A & N/A & N/A & N/A & N/A & N/A & N/A & N/A & 12.9ms* & 12.9ms \\
    \hline
    NLM Path & 1.2s & 0s & 0s & 4.7s & 0s & 0s & 3.6s & 1.5s & 7.3s & 18.3s\\
    \hline
    NLM Sort & 2.6s & 0s & 0s & 11.1s & 0s & 0s & 7.5s & 3.4s & 17.1s & 41.7s\\
    \hline
    NLM Blocks World & 635ms & 0ms & 0ms & 2100ms & 0ms & 0ms & 1400ms & 618ms & 3100ms & 7850ms\\
    \hline
\end{tabular}
\label{tab:submodel_runtimes}
\end{table*}

\begin{figure*}[t]
\centerline{\includegraphics[width = \textwidth]{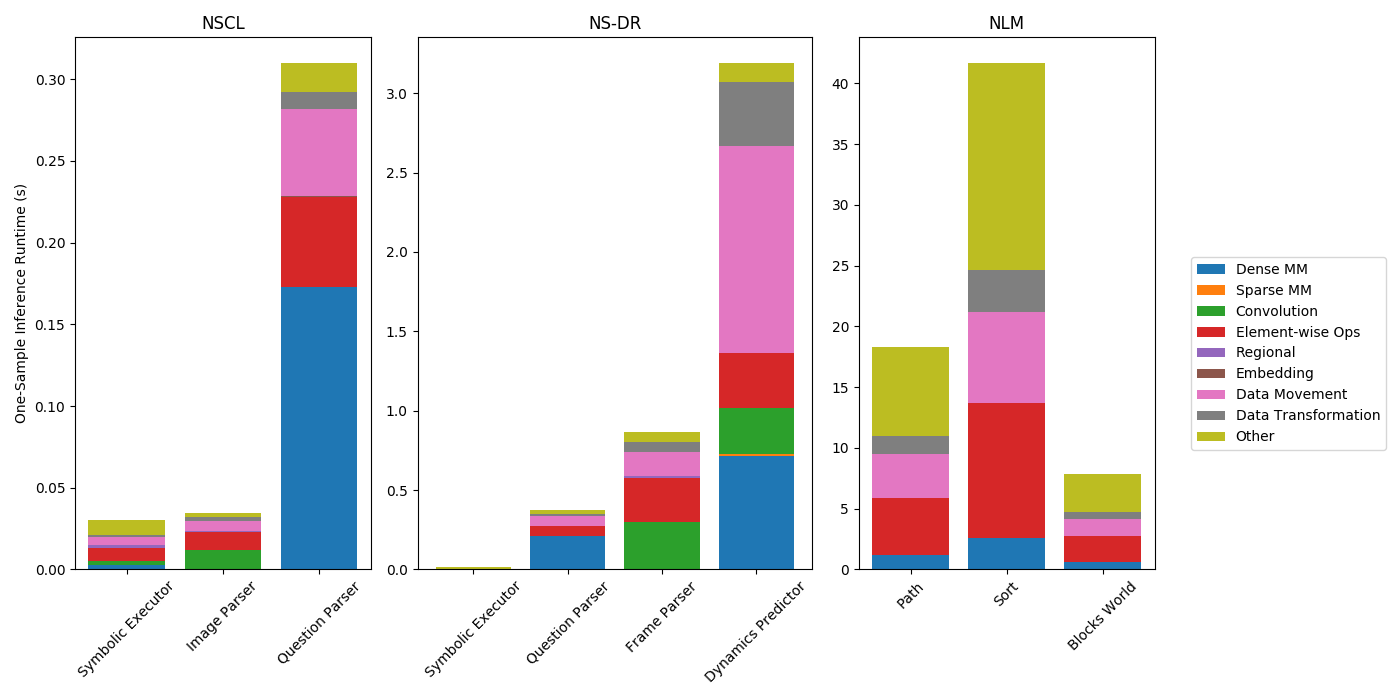}}
\caption{Runtime breakdown for each of the three models, including submodels for NSCL and NS-DR and three distinct tasks for NLM. Note the differing Y scales for the three models.}
\label{fig:model_runtimes}
\end{figure*}

In this section, we present and discuss results for the individual submodels introduced in Section \ref{methodology}. We first present and discuss the performance characterizations of the individual submodels, then discuss the aggregate inference runtime behavior of the NSCL, NS-DR, and NLM models. 
    
\subsection{Image and Video Frame Parser}

Mask R-CNN, the video frame parser, has a well-studied performance profile that is dominated by convolution and activation functions \cite{he2018mask}. As mentioned in Section \ref{model_overview}, we used Detectron2 due to difficulties bringing up the version of Mask R-CNN used in the NS-DR paper. Mobile-optimized, production ready implementations of Detectron2 models are provided by the d2go project \cite{d2go}, which we used to collect realistic performance measurements for Mask R-CNN inference latency. The pretrained model we used was \textit{not} trained on the CLEVR or CLEVRER dataset, as there is no pretrained model, model source, or training instructions for CLEVRER, and we wanted to use the same submodel for both the NSCL and NS-DR. However, we do not anticipate that the performance characteristics of the model would be any different if it were trained for a different dataset, particularly since the runtime of this model does not significantly vary with the number of objects in the input image.

Our analysis of Detectron2, as shown in Table \ref{tab:submodel_runtimes} and Figure \ref{fig:model_runtimes}, shows that Mask R-CNN spends the most execution time on convolution and element-wise operations (such as activation functions and normalization). This is not particularly surprising, but serves as a simple example of our classification scheme. With an average 34.6ms inference time on our target machine, a full 25-frame video requires 865ms of inference time for the CLEVRER dataset. Thus, the NS-DR video frame parser takes dramatically longer than the NSCL image parser.


\subsection{Question Parser}

NLP models such as Seq2Seq, the original question parser in the NS-DR, are a well-established family, having seen datacenter deployments since at least 2016\cite{sutskever2014sequence}. However, transformer-based attention models, such as the OpenNMT model we profiled, typically outperform RNN-based models such as Seq2Seq, and have seen wider adoption since 2017\cite{vaswani2017attention}.\par

The amount of computation required to perform inference for an NMT model is dependent on the input sequence length. Sentences in the CLEVR dataset have an average length of 18.4 words. The CLEVRER dataset is split between open-ended questions with an average length of 10.9 words and multiple-choice questions with an average length of 51.3 words (counting the answer choices), for an overall average of 22.2 words.\par

Figure \ref{fig:model_runtimes} shows that the performance of the transformer-based OpenNMT model is dominated by dense matrix multiplication and data movement. The OpenNMT model performed inference in an average of 13.5ms per input word/token. In practice, the runtime of a transformer model asymptotically grows quadratically with input sequence length\cite{transformer_quadratic}, but we observed a linear relation on sample input sentences. It is likely that, for the relatively short lengths of the inputs we were observing, linear per-word operations dominated the quadratic attention operations.\par

\subsection{Dynamics Predictor}

PropNet, the neural dynamics predictor, spends the majority of its runtime on data movement. In fact, while analyzing the behavior of this model using the \texttt{nvidia-smi} utility, we noticed it was rarely able to exceed 50\% utilization of one GPU. The model also spends a substantial amount of time on coalescing, which entails merging duplicate entries in sparse tensors. It is unclear to what extent this coalescing could be avoided through better-optimized code. A substantial portion of the remaining runtime is spent on dense matrix multiplication. This is because the workload involves the multiplication of many very tall and skinny matrices, where $m$ and $n$ are quite large, but $k$ may be as small as 1. As discussed in Section \ref{methodology}, these types of matrices are challenging to multiply efficiently.\par

Internally, this model consists of several smaller fully-connected and convolutional networks. Some of these models must be run multiple times for a single frame in order to compute the propagation of forces. This contributes to this model's large amount of data movement and generally long runtime. The data movement of this model could be reduced by leveraging sparsity and weight reuse; since the same submodel is used for inference many times, it is wasteful to move over the weights for each inference.\par

Some tensor dimensions are input-dependent, corresponding to the number of objects in the frame. We also observed substantial variation in inference runtimes. Across 400 samples, the fastest inference finished in 1.8 seconds, and the slowest in 5.3 seconds, with an average of 3.4 seconds.\par


\subsection{NSCL Symbolic Program Executor}
The symbolic executor for the NSCL spends a large portion of its execution time on element-wise operations, and a smaller but still significant amount of time on data movement. These element-wise operations stem from its manipulation of vectors of probabilities (with entries corresponding to the probabilities of each object in the scene being a correct answer). In addition to simple arithmetic, the model computes numerous softmax functions over these vectors in order to isolate predictions. Notably, not much time is spent on matrix-matrix operations (GEMM and convolution), which would have better operational intensity and potential for parallelism than the scalar and vector operations we see dominating.

\subsection{NS-DR Symbolic Program Executor}

\begin{figure}[htbp]
\centerline{\includegraphics[width = 0.9\columnwidth]{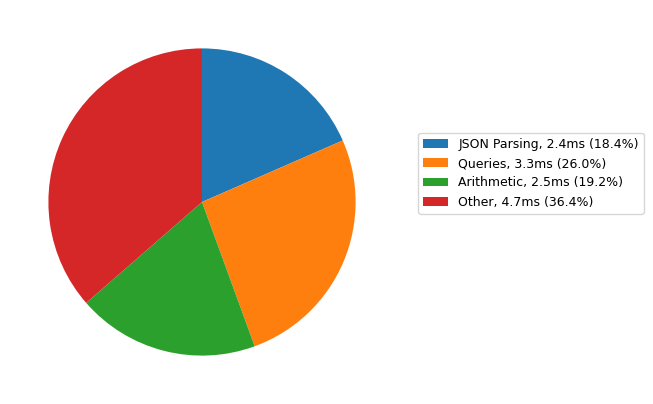}}
\caption{Breakdown of the NS-DR's symbolic component (This model is broken down separately since it is the only CPU-only component.)}
\label{fig:executor}
\end{figure}

The symbolic program executor for the NS-DR has an average runtime of 12.9ms per sample. While this is not large compared to the other models in the network, all other models must finish before the executor can begin, and it therefore lies on the critical path for inference.\par

Since the NS-DR executor is a CPU-only, scalar model, it would not make sense to categorize it according to the scheme we derived for the other workloads. Therefore, we used Python's built-in cProfile profiler to perform function-level profiling and identified categories of similar functions. The results of this analysis are shown in Figure \ref{fig:executor}. Note, however, that the plurality of the runtime still falls into the "Other" category. This is a consequence of there being many miscellaneous functions which individually contribute little to runtime, but collectively contribute more than any of the major categories.

Much of the executor's runtime is spent on querying the set of extracted features. Examples of these queries include finding the ID of an object with given properties, looking up the properties of an object given its ID, and queries relating multiple objects such as finding an object's closest neighbor. These queries look similar to standard database operations. For a sufficiently large set of symbolic features, it is likely that feature querying could be approached using existing work in parallelizing SQL queries \cite{parallel_sql}. However, the feature set the NS-DR extracts is relatively small, meaning that the overhead which parallelism incurs would almost certainly overwhelm the speedup for this model.\par

The next-largest category of execution time is spent on scalar arithmetic operations, in particular summing large arrays. In principle these sorts of operations can sometimes benefit from being spread across multiple CPU cores, but in practice we again expect the small size of the feature set to make this pointless.\par

The third-largest category of operation is in fact JSON parsing. This is an unfortunate artifact of the way data is passed in the NS-DR model: rather than an end-to-end integration, it stores data in JSON files between submodels. This means that the executor has to load in the questions and extracted features from JSONs for each inference sample. This overhead turns out to be substantial for the very short runtime of this model.\par



\subsection{Neural Logic Machines}
A characterization profile was created for each of the NLM tasks: Sort, Path, and Blocks World. Each task provided ranging variations in the produced results. The largest category of operation was the "other" category, where a significant amount of overall time was spent. The percentage of time spent in the other category can be determined as 40.9 percent for the Sort task, 40.0 percent for the critical path task, and 45.1 percent for the Blocks World task. The amount of time spent in the other category can be attributed to miscellaneous internal querying and generation functions for the different tasks.

Following the "other" category, the element-wise category contributed the second most to the task executions. The Sort task resulted in the highest percentage of element-wise functions related to the vector multiplication being performed on the numerical task. The high percentage of element-wise functions is also represented in the Path task, with approximately a third of the total runtime spent on element-wise tasks.

Data movement represents the third largest category that was characterized during the task executions. The Sort task and Path task shared the highest percentage of data movement within the model's execution. This is predominantly due to the frequent movement of numerical values in these general tasks. Blocks World yielded the lowest data movement characteristic which can be perceived by the more symbolic and logical flow of functions and it's data. 

\section{Analysis}
\label{analysis}

Figure \ref{fig:model_runtimes} shows the execution time breakdowns for single input samples for each of the three models in this paper. We now discuss each model individually.

\subsection{NSCL and NS-DR}
The NSCL runs in an average of 375ms per sample, with 310ms of this coming just from the question parser. The clear bottleneck for this workload, dominating the question parser's runtime, is dense matrix multiplication. This is another case where a sparse version of the model could potentially yield improvements to runtime. However, this is not the only contributing factor to the question parser's long runtime. Although the ability to ask questions of a model using natural language is convenient, it is worth keeping in mind that it can add substantial overhead and complexity, and considering whether a simpler, machine-readable input "language" could be used instead.\par

It is clear that the dynamics predictor dominates the inference runtime of the NS-DR. The total inference time for this model is 4.44 seconds. Since input videos are 5 seconds long, this does in theory allow for real-time inference on this system, but this would not hold true for low-power or embedded platforms.\par

Some portions of the NS-DR and NSCL are more suited for acceleration than others. The CNN-based video frame parser and the transformer-based question parser model can be accelerated with existing GPU or accelerator architectures \cite{centuar_accelerator}. On the other hand, the NS-DR's dynamics predictor is severely data-movement-bound due to its many sparse and small tensor operations, and also suffers from the difficulty of multiplying "tall-and-skinny" matrices. It is possible that data movement could be reduced somewhat through code optimization, but it is fundamentally difficult to accelerate workloads with poor operational intensity. There have also been several approaches investigated for near-memory linear algebra \cite{near_memory_sparse}\cite{spacea}, which would likely be beneficial for PropNet as the reduced time accessing memory would help compensate for the poor operational intensity.\par

The symbolic executor of the NS-DR exhibits complex control flow and heterogeneous functionality. Overall, this workload has very little of the coarse-grain parallelism typical of deep learning workloads. However, there are still opportunities for fine-grain parallelism in its database queries and arithmetic operations. It is doubtful that a multicore CPU implementation would give any speedup for this particular workload, given the relatively small sizes of the feature sets, but this could be a viable approach for larger symbolic models.\par

Our biggest takeaway from analyzing the performance of the NSCL and NS-DR models is that these NSAI models still perform largely the same work as other AI models that have been previously studied, in large part due to the small amount of time spent on symbolic execution. We were able to largely classify the types of operations performed by the neural dynamics predictor, the bottleneck in the NS-DR, into the same primitive operations performed by other ML models in different domains. Although the symbolic executor is currently a small fraction of the inference runtime for both models, it is likely that this component's complexity will increase over time as model developers seek to extract and manipulate more complex relations between features. We believe that it is possible to leverage work from similar program architectures, such as relational databases, to reduce the impact of this.

\subsection{NLM}
The NLM showcased the ability for the architecture to train a model on small scale tasks and generalize to solve large scale tasks with lifted rules and added premises, which shows the expressive power of the network. The ability to scale the rule set from a small sized rule set to a large sized rule set has proven to be difficult for inductive logic programming systems. The combination of using both symbols and probabilities helps solve these issues. The NLM proves to accomplish complex tasks by overcoming major challenges that traditional neural networks and inductive logic reasoning systems cannot solve alone. The tasks that were used involved the graph-based path task, the general sort application task, and the more complex Blocks World task. These different problems presented variations in the system's performance, as each task involved different levels of logic rule set complexity. In addition, the NLM architecture also solves the problem of scalability with respect to the complexity of rules given to the system. As the rules of the task scale up, the complexity of the logic rules to be learned will also scale up exponentially. This allows the NLM to adjust its trained rules based on uses of a minimal set of prior examples. Using a minimum set of prior data illustrates the ability for the NLM to effectively improve as it learns. As these different tasks use a variation of input data and parameters, the performance evaluation of these exercises gives many opportunities for improvement.

\section{Conclusion}
\label{conclusion}
%
%
%
While neuro-symbolic models with submodels look topologically distinct from traditional deep learning models, our analysis suggests that their performance characteristics can be largely viewed as a combination of existing workloads. Our analysis of the NSCL and NS-DR show that there are relatively few opportunities for acceleration of symbolic computation. The symbolic workloads of these two models have low operational intensities, consisting of vector and/or scalar operations, and exhibit complex control flow. These factors combined greatly limit the potential for parallelism. However, the symbolic components do not make up large portions of the execution times of either workload, and are therefore unlikely to pose a bottleneck. On the other hand, the NLP, dynamics predictor, and vision submodels exhibit different dominant operation categories, including dense GEMM in the question parser, data movement in the dynamics predictor, and a co-dominance of convolution and element-wise operations in the image and frame parser. NLM models require numerous element-wise operations for inference, with the computational demand increasing with the complexity of the task-specific logical rule set.\par

The challenge of accelerating the low-operational-intensity, element-wise computations of workloads such as NLM will become increasingly important as this field sees further development.

\section*{Acknowledgement}
\noindent
This research was supported in part by Semiconductor Research Corporation (SRC) Task 3015.001/3016.001 and National Science Foundation grant number 1763848. Any opinions, findings, conclusions or recommendations are those of the authors and not of the funding agencies.

\appendix
Repository links for the models referenced in this paper:
\resizebox{0.95\columnwidth}{!}{\begin{tabular}{|l|l|} 
\hline
\textbf{NSCL} & https://github.com/vacancy/NSCL-PyTorch-Release\\
\hline
\textbf{NS-DR} & https://github.com/chuangg/CLEVRER\\
\hline
\textbf{Detectron2 (d2go)} & https://github.com/facebookresearch/d2go\\
\hline
\textbf{OpenNMT} & https://github.com/OpenNMT/OpenNMT-py\\
\hline
\textbf{Propnet} & https://github.com/YunzhuLi/PropNet\\
\hline
\textbf{NLM} & https://github.com/google/neural-logic-machines\\
\hline
\end{tabular}}

\bibliographystyle{IEEEtran}
\bibliography{bibliography}

\begin{thebibliography}{10}
\providecommand{\url}[1]{#1}
\csname url@samestyle\endcsname
\providecommand{\newblock}{\relax}
\providecommand{\bibinfo}[2]{#2}
\providecommand{\BIBentrySTDinterwordspacing}{\spaceskip=0pt\relax}
\providecommand{\BIBentryALTinterwordstretchfactor}{4}
\providecommand{\BIBentryALTinterwordspacing}{\spaceskip=\fontdimen2\font plus
\BIBentryALTinterwordstretchfactor\fontdimen3\font minus
  \fontdimen4\font\relax}
\providecommand{\BIBforeignlanguage}[2]{{%
\expandafter\ifx\csname l@#1\endcsname\relax
\typeout{** WARNING: IEEEtran.bst: No hyphenation pattern has been}%
\typeout{** loaded for the language `#1'. Using the pattern for}%
\typeout{** the default language instead.}%
\else
\language=\csname l@#1\endcsname
\fi
#2}}
\providecommand{\BIBdecl}{\relax}
\BIBdecl

\bibitem{MONTAVON20181}
\BIBentryALTinterwordspacing
G.~Montavon, W.~Samek, and K.-R. Müller, ``Methods for interpreting and
  understanding deep neural networks,'' \emph{Digital Signal Processing},
  vol.~73, pp. 1--15, 2018. [Online]. Available:
  \url{https://www.sciencedirect.com/science/article/pii/S1051200417302385}
\BIBentrySTDinterwordspacing

\bibitem{GARNELO201917}
\BIBentryALTinterwordspacing
M.~Garnelo and M.~Shanahan, ``Reconciling deep learning with symbolic
  artificial intelligence: representing objects and relations,'' \emph{Current
  Opinion in Behavioral Sciences}, vol.~29, pp. 17--23, 2019, artificial
  Intelligence. [Online]. Available:
  \url{https://www.sciencedirect.com/science/article/pii/S2352154618301943}
\BIBentrySTDinterwordspacing

\bibitem{lighthill}
J.~Lighthill, ``Artificial intelligence: A general survey,'' in
  \emph{Artificial Intelligence: a paper symposium}, 1973.

\bibitem{hendler2008avoiding}
J.~Hendler, ``Avoiding another ai winter,'' \emph{IEEE Annals of the History of
  Computing}, vol.~23, no.~02, pp. 2--4, 2008.

\bibitem{mao2019neurosymbolic}
J.~Mao, C.~Gan, P.~Kohli, J.~B. Tenenbaum, and J.~Wu, ``The neuro-symbolic
  concept learner: Interpreting scenes, words, and sentences from natural
  supervision,'' 2019.

\bibitem{tbd_model}
\BIBentryALTinterwordspacing
D.~Mascharka, P.~Tran, R.~Soklaski, and A.~Majumdar, ``Transparency by design:
  Closing the gap between performance and interpretability in visual
  reasoning,'' \emph{CoRR}, vol. abs/1803.05268, 2018. [Online]. Available:
  \url{http://arxiv.org/abs/1803.05268}
\BIBentrySTDinterwordspacing

\bibitem{mac_model}
\BIBentryALTinterwordspacing
D.~A. Hudson and C.~D. Manning, ``Compositional attention networks for machine
  reasoning,'' in \emph{International Conference on Learning Representations},
  2018. [Online]. Available: \url{https://openreview.net/forum?id=S1Euwz-Rb}
\BIBentrySTDinterwordspacing

\bibitem{yi2020clevrer}
K.~Yi, C.~Gan, Y.~Li, P.~Kohli, J.~Wu, A.~Torralba, and J.~B. Tenenbaum,
  ``Clevrer: Collision events for video representation and reasoning,'' 2020.

\bibitem{neural_logic_machines}
\BIBentryALTinterwordspacing
H.~Dong, J.~Mao, T.~Lin, C.~Wang, L.~Li, and D.~Zhou, ``Neural logic
  machines,'' \emph{CoRR}, vol. abs/1904.11694, 2019. [Online]. Available:
  \url{http://arxiv.org/abs/1904.11694}
\BIBentrySTDinterwordspacing

\bibitem{han2020visual}
C.~Han, J.~Mao, C.~Gan, J.~B. Tenenbaum, and J.~Wu, ``Visual
  concept-metaconcept learning,'' 2020.

\bibitem{zfchen2021iclr}
Z.~Chen, J.~Mao, J.~Wu, K.-Y.~K. Wong, J.~B. Tenenbaum, and C.~Gan, ``Grounding
  physical concepts of objects and events through dynamic visual reasoning,''
  in \emph{International Conference on Learning Representations}, 2021.

\bibitem{blocks_world}
\BIBentryALTinterwordspacing
N.~J. Nilsson, ``Principles of artificial intelligence,'' in \emph{Principles
  of Artificial Intelligence}.\hskip 1em plus 0.5em minus 0.4em\relax
  Springer-Verlag Berlin Heidelberg, 1982, pp. 1--476. [Online]. Available:
  \url{https://www.springer.com/gp/book/9783540113409}
\BIBentrySTDinterwordspacing

\bibitem{mao2018the}
\BIBentryALTinterwordspacing
J.~Mao, C.~Gan, P.~Kohli, J.~B. Tenenbaum, and J.~Wu, ``The neuro-symbolic
  concept learner: Interpreting scenes, words, and sentences from natural
  supervision,'' in \emph{International Conference on Learning
  Representations}, 2019. [Online]. Available:
  \url{https://openreview.net/forum?id=rJgMlhRctm}
\BIBentrySTDinterwordspacing

\bibitem{low_latency_rnns}
\BIBentryALTinterwordspacing
P.~Gao, L.~Yu, Y.~Wu, and J.~Li, ``Low latency rnn inference with cellular
  batching,'' in \emph{Proceedings of the Thirteenth EuroSys Conference}, ser.
  EuroSys '18.\hskip 1em plus 0.5em minus 0.4em\relax New York, NY, USA:
  Association for Computing Machinery, 2018. [Online]. Available:
  \url{https://doi.org/10.1145/3190508.3190541}
\BIBentrySTDinterwordspacing

\bibitem{lightweight_mask_rcnn}
\BIBentryALTinterwordspacing
H.~Li, A.~Wu, W.~Fang, Q.~Zhang, M.~Liu, Q.~Liu, and W.~Chen, ``Lightweight
  mask {R-CNN} for long-range wireless power transfer systems,'' \emph{CoRR},
  vol. abs/2004.08761, 2020. [Online]. Available:
  \url{https://arxiv.org/abs/2004.08761}
\BIBentrySTDinterwordspacing

\bibitem{wu2019detectron2}
Y.~Wu, A.~Kirillov, F.~Massa, W.-Y. Lo, and R.~Girshick, ``Detectron2,''
  \url{https://github.com/facebookresearch/detectron2}, 2019.

\bibitem{klein2017opennmt}
G.~Klein, Y.~Kim, Y.~Deng, J.~Senellart, and A.~M. Rush, ``Opennmt: Open-source
  toolkit for neural machine translation,'' 2017.

\bibitem{li2018propagation}
Y.~Li, J.~Wu, J.-Y. Zhu, J.~B. Tenenbaum, A.~Torralba, and R.~Tedrake,
  ``Propagation networks for model-based control under partial observation,''
  in \emph{ICRA}, 2019.

\bibitem{clevr_datset}
\BIBentryALTinterwordspacing
J.~Johnson, B.~Hariharan, L.~van~der Maaten, L.~Fei{-}Fei, C.~L. Zitnick, and
  R.~B. Girshick, ``{CLEVR:} {A} diagnostic dataset for compositional language
  and elementary visual reasoning,'' \emph{CoRR}, vol. abs/1612.06890, 2016.
  [Online]. Available: \url{http://arxiv.org/abs/1612.06890}
\BIBentrySTDinterwordspacing

\bibitem{he2018mask}
K.~He, G.~Gkioxari, P.~Dollár, and R.~Girshick, ``Mask r-cnn,'' 2018.

\bibitem{cho-etal-2014-learning}
\BIBentryALTinterwordspacing
K.~Cho, B.~van Merri{\"e}nboer, C.~Gulcehre, D.~Bahdanau, F.~Bougares,
  H.~Schwenk, and Y.~Bengio, ``Learning phrase representations using {RNN}
  encoder{--}decoder for statistical machine translation,'' in
  \emph{Proceedings of the 2014 Conference on Empirical Methods in Natural
  Language Processing ({EMNLP})}.\hskip 1em plus 0.5em minus 0.4em\relax Doha,
  Qatar: Association for Computational Linguistics, Oct. 2014, pp. 1724--1734.
  [Online]. Available: \url{https://aclanthology.org/D14-1179}
\BIBentrySTDinterwordspacing

\bibitem{bahdanau2016neural}
D.~Bahdanau, K.~Cho, and Y.~Bengio, ``Neural machine translation by jointly
  learning to align and translate,'' 2016.

\bibitem{berkeley_dwarfs}
K.~Asanovic, R.~Bodik, B.~Catanzaro, J.~Gebis, P.~Husbands, K.~Keutzer,
  D.~Patterson, W.~Plishker, J.~Shalf, S.~Williams, and K.~Yelick, ``The
  landscape of parallel computing research: A view from berkeley,'' 12 2006.

\bibitem{Rivera_2021}
\BIBentryALTinterwordspacing
C.~Rivera, J.~Chen, N.~Xiong, J.~Zhang, S.~L. Song, and D.~Tao, ``Tsm2x:
  High-performance tall-and-skinny matrix–matrix multiplication on gpus,''
  \emph{Journal of Parallel and Distributed Computing}, vol. 151, p. 70–85,
  May 2021. [Online]. Available:
  \url{http://dx.doi.org/10.1016/j.jpdc.2021.02.013}
\BIBentrySTDinterwordspacing

\bibitem{structured_sparsity}
\BIBentryALTinterwordspacing
A.~Zhou, Y.~Ma, J.~Zhu, J.~Liu, Z.~Zhang, K.~Yuan, W.~Sun, and H.~Li,
  ``Learning {N:} {M} fine-grained structured sparse neural networks from
  scratch,'' \emph{CoRR}, vol. abs/2102.04010, 2021. [Online]. Available:
  \url{https://arxiv.org/abs/2102.04010}
\BIBentrySTDinterwordspacing

\bibitem{fast_conv}
A.~Vasudevan, A.~Anderson, and D.~Gregg, ``Parallel multi channel convolution
  using general matrix multiplication,'' in \emph{2017 IEEE 28th International
  Conference on Application-specific Systems, Architectures and Processors
  (ASAP)}, 2017, pp. 19--24.

\bibitem{embedding_acceleration}
L.~Ke, U.~Gupta, B.~Y. Cho, D.~Brooks, V.~Chandra, U.~Diril, A.~Firoozshahian,
  K.~Hazelwood, B.~Jia, H.-H.~S. Lee, M.~Li, B.~Maher, D.~Mudigere, M.~Naumov,
  M.~Schatz, M.~Smelyanskiy, X.~Wang, B.~Reagen, C.-J. Wu, M.~Hempstead, and
  X.~Zhang, ``Recnmp: Accelerating personalized recommendation with near-memory
  processing,'' in \emph{2020 ACM/IEEE 47th Annual International Symposium on
  Computer Architecture (ISCA)}, 2020, pp. 790--803.

\bibitem{pytorch_sparse_doc}
\BIBentryALTinterwordspacing
``torch.sparse.'' [Online]. Available:
  \url{https://pytorch.org/docs/stable/sparse.html}
\BIBentrySTDinterwordspacing

\bibitem{d2go}
``D2go brings detectron2 to mobile,''
  \url{https://ai.facebook.com/blog/d2go-brings-detectron2-to-mobile/},
  accessed: 2021-05-09.

\bibitem{sutskever2014sequence}
I.~Sutskever, O.~Vinyals, and Q.~V. Le, ``Sequence to sequence learning with
  neural networks,'' 2014.

\bibitem{vaswani2017attention}
A.~Vaswani, N.~Shazeer, N.~Parmar, J.~Uszkoreit, L.~Jones, A.~N. Gomez,
  L.~Kaiser, and I.~Polosukhin, ``Attention is all you need,'' 2017.

\bibitem{transformer_quadratic}
\BIBentryALTinterwordspacing
A.~Katharopoulos, A.~Vyas, N.~Pappas, and F.~Fleuret, ``Transformers are rnns:
  Fast autoregressive transformers with linear attention,'' \emph{CoRR}, vol.
  abs/2006.16236, 2020. [Online]. Available:
  \url{https://arxiv.org/abs/2006.16236}
\BIBentrySTDinterwordspacing

\bibitem{parallel_sql}
\BIBentryALTinterwordspacing
T.~Cruanes, B.~Dageville, and B.~Ghosh, ``Parallel sql execution in oracle
  10g,'' in \emph{Proceedings of the 2004 ACM SIGMOD International Conference
  on Management of Data}, ser. SIGMOD '04.\hskip 1em plus 0.5em minus
  0.4em\relax New York, NY, USA: Association for Computing Machinery, 2004, p.
  850–854. [Online]. Available: \url{https://doi.org/10.1145/1007568.1007666}
\BIBentrySTDinterwordspacing

\bibitem{centuar_accelerator}
G.~Henry, P.~Palangpour, M.~Thomson, J.~S. Gardner, B.~Arden, J.~Donahue,
  K.~Houck, J.~Johnson, K.~O’Brien, S.~Petersen, B.~Seroussi, and T.~Walker,
  ``High-performance deep-learning coprocessor integrated into x86 soc with
  server-class cpus industrial product,'' in \emph{2020 ACM/IEEE 47th Annual
  International Symposium on Computer Architecture (ISCA)}, 2020, pp. 15--26.

\bibitem{near_memory_sparse}
\BIBentryALTinterwordspacing
D.~Fujiki, N.~Chatterjee, D.~Lee, and M.~O'Connor, ``Near-memory data
  transformation for efficient sparse matrix multi-vector multiplication,'' in
  \emph{Proceedings of the International Conference for High Performance
  Computing, Networking, Storage and Analysis}, ser. SC '19.\hskip 1em plus
  0.5em minus 0.4em\relax New York, NY, USA: Association for Computing
  Machinery, 2019. [Online]. Available:
  \url{https://doi.org/10.1145/3295500.3356154}
\BIBentrySTDinterwordspacing

\bibitem{spacea}
X.~Xie, Z.~Liang, P.~Gu, A.~Basak, L.~Deng, L.~Liang, X.~Hu, and Y.~Xie,
  ``Spacea: Sparse matrix vector multiplication on processing-in-memory
  accelerator,'' in \emph{2021 IEEE International Symposium on High-Performance
  Computer Architecture (HPCA)}, 2021, pp. 570--583.

\end{thebibliography}

\end{document}